\documentclass{l4dc2027}
\usepackage{enumitem}
\usepackage{times}
\usepackage{mathtools}
\usepackage{booktabs}
\usepackage{appendix}

\newtheorem{assumption}[theorem]{Assumption}

\title{Cancellation-Free Constrained SquareCB for Safe Controller Selection}

\author{%
 \Name{Kartik Pandit} \Email{ksp82@njit.edu}
 \AND
 \Name{Sourav Ganguly} \Email{sg2786@njit.edu}
 \AND
 \Name{Arnob Ghosh} \Email{arnob.ghosh@njit.edu}\\
 \addr Electrical and Computer Engineering Department, NJIT, Newark, NJ, USA
}

\begin{document}

\maketitle

\vspace{-0.2in}
\begin{abstract}%
We study constrained contextual bandits, in which a learner chooses one of $K$ actions in
each round to minimize loss while keeping the expected cost of its play below a budget.
Existing reductions to online regression weight the cost by a Lagrange multiplier learned from past constraint residuals, so slack in some rounds can offset violations in others. We instead run SquareCB proposed in \citep{foster2020beyond} with a rectified weight on the cost: it is zero when the estimated cost is within the budget, and otherwise just large enough to bring it back to the budget. Under realizability and Slater's condition with margin $\delta$, the algorithm
attains regret $O(\sqrt{KT\,\mathrm{Reg}_{\mathsf{Sq}}(T)}/\delta)$ and cancellation-free
violation $\sum_t(V_c^{p_t}-b)^+=O(\sqrt{KT\,\mathrm{Reg}_{\mathsf{Sq}}(T)})$, where
$\mathrm{Reg}_{\mathsf{Sq}}$ is the error of the regression oracle. The same bounds hold when
the budget must be met in every context, potentially adversarially chosen contexts. In controller
selection for a safe-control-gym quadrotor and a BOPTEST building, the violation is 6 to 9
times lower than that of primal-dual and virtual-queue baselines, with a lower or comparable regret.%
\end{abstract}

\vspace{-0.2in}
\section{Introduction}
\label{sec:intro}
\vspace{-0.105in}
Contextual bandits (CBs) model sequential decisions under uncertainty: in each round the
learner observes a context, chooses an action, and observes feedback for that action only
\citep{langford2007epoch, lattimore2020bandit}. They are used in recommendation
\citep{li2010contextual}, crowdsourcing \citep{muller2018context} and medicine
\citep{tewari2017ads}. Many such problems also carry an operational requirement that the
objective alone does not capture: an advertising budget, a fairness rule, or a safety limit.
Constrained contextual bandits (CCBs) add a cost to each action and ask the learner to
minimize loss while keeping the expected cost below a budget $b$
\citep{
slivkins2023contextual, guo2024stochastic, sarkar2026simple}.
\vspace{-0.1in}
\paragraph{Controller selection.} CCBs arise naturally in control when a supervisor selects
among a library of pre-designed controllers \citep{ayala2024risk,pacchiano2020model}. Before each episode, the supervisor observes the
operating conditions (a payload, a start state, a weather forecast), selects a controller, runs
it for the episode, and observes a performance loss and a safety or comfort cost.  The supervisor has to learn which controller to use in which conditions
while keeping the cost within a budget, and it should not repeatedly exceed the budget early
in the run and compensate later.
\vspace{-0.1in}
\paragraph{Regression oracles and signed guarantees.} \citet{foster2020beyond} reduced
contextual bandits to online square-loss regression: SquareCB plays inverse-gap weighting on
the predictions of any online regression oracle, and its regret depends on the class only
through the oracle's error. For constraints, \citet{slivkins2023contextual} combine SquareCB
with a Lagrange multiplier updated by online learning, and \citet{sarkar2026simple} replace
the multiplier with a virtual queue and a Lyapunov function, which also handles adversarial
contexts. In both, the weight on the cost is driven by the signed residual $V_c^{p_t}-b$, and
the long-term guarantee bounds the signed sum $\sum_t(V_c^{p_t}-b)$. This sum allows
cancellation: slack banked in rounds below the budget offsets violations in later rounds. If
$V_c^{p_t}-b$ alternates between $+\epsilon$ and $-\epsilon$, the  sum is at most
$\epsilon$, while the budget is exceeded in half of the rounds. On the other hand, cancellation-free bound is
$\sum_t(V_c^{p_t}-b)^+=\epsilon T/2$, and thus, it is a  stronger metric. Further, they did not consider the cancellation free regret bound $\sum_t(V_r^{p_t}-V_r^{\pi^*})^+$, hence, the regret can be made small by playing infeasible but high reward policies.
\vspace{-0.1in}
\paragraph{Our approach.} We keep SquareCB and change only how the cost weight is chosen. In
each round, the weight $\mu_t\in[0,\Lambda]$ is the smallest value for which the estimated
cost of the inverse-gap-weighted distribution on the scores
$\hat f_r^t+\mu_t\hat f_c^t$ is at most $b$, found by bisection; if the estimated cost jumps over
$b$, the learner mixes the two neighbouring distributions so that it equals $b$. The weight is
therefore a function of the current estimates alone. In every round, either the weight is $0$
and the estimated cost is below the budget, or the estimated cost equals the budget, or the
weight is at its cap $\Lambda$. This per-round complementary slackness, combined with an
exact-penalty saddle inequality, turns the SquareCB analysis into a bound on the one-sided
violation. Using the one-sided violation, we also bound the cancellation-free regret. 
To our knowledge, the cancellation-free
regret has not been bounded before for constrained contextual bandits with regression oracles, and the cancellation-free violation appears only in concurrent work\footnote{Concurrently with and independently of our work, the October 2026 version of
\citep{sarkar2026constrained} also the cancellation-free violation. However, the inherent technique is very different from ours. Further, unlike our work, their work does not achieve a cancellation-free regret bounds.}.

\textbf{Contributions}: Our major contributions are listed below
\vspace{-0.1in}
\begin{itemize}[leftmargin=*]
\item We develop a novel  algorithm for constrained contextual bandits with general online regression oracles,
  which adds a bisection on a single scalar per round to SquareCB and has no dual-ascent loop
  (Section~\ref{sec:setup}--Proposition~\ref{prop:fp}). Under realizability and Slater's condition with margin $\delta$, and with
  $\rho_T=\sqrt{KT\,\mathrm{Reg}_{\mathsf{Sq}}(T)}$: regret $O(\rho_T/\delta)$ and
  cancellation-free violation $O(\rho_T)$, which does not depend on $\delta$
  (Theorems~\ref{thm:viol} and~\ref{thm:reward}). The cancellation-free regret
  $\sum_t(V_r^{p_t}-V_r^{\pi^*})^+$ is also $O(\rho_T/\delta)$ (Corollary~\ref{cor:posreg}).\vspace{-0.05in}
\item The same rates apply when the budget must hold in every context, for arbitrary, possibly
  adversarial, context sequences (Theorem~\ref{thm:ctx}). \vspace{-0.05in}
\item Experiments on controller selection for a safe-control-gym quadrotor and a BOPTEST
  building, in which the violation is 6 to 9 times lower than that of primal-dual and
  virtual-queue baselines, with lower regret (Section~\ref{sec:results}).
\end{itemize}


\vspace{-0.1in}
\subsection{Related Work}\label{sec:related}

\vspace{-0.03in}
Contextual bandits have been studied with linear
\citep{chu2011contextual,abbasi2011improved,agrawal2013thompson} and neural
\citep{zahavy2019deep,zhou2020neural,zhang2020neural} function classes using optimism on the face of uncertainty principle.
 An alternative to optimism is to design algorithms for specific function classes by reducing contextual bandit learning to online regression. In particular, the Inverse Gap Weighting (IGW) principle \citep{abe1999associative,abe2003reinforcement} underlies general regression-oracle-based reductions developed by \citep{foster2020beyond,foster2018practical} (e.g., \texttt{Square-CB}) which decouple exploration from the choice of function approximator. \citet{deb2025conservative}
apply this to conservative bandits, which require performance relative to a baseline policy
rather than a cost budget. 

Unlike conservative contextual bandits, which ensure performance relative to a baseline policy, constrained contextual bandits explicitly optimize a reward objective subject to a cost budget.  For constrained contextual bandits with regression oracles,
\citet{slivkins2023contextual} use Lagrange multipliers and \citet{sarkar2026simple} use virtual
queues, which also handle adversarial contexts. Both drive the cost weight by the signed
residual and bound the signed violation, in which rounds below the budget offset rounds above
it. Applying them to the clipped cost $(c-b)^+$ removes the cancellation but bounds but bounds
$\sum_t\mathbb{E}_{a\sim p_t}(c(a)-b)^+\ge\sum_t(V_c^{p_t}-b)^+$, a different pessimistic constraint. Positive-part guarantees have been studied in constrained MDPs \citep{efroni2020exploration,guo2022online,muller2024truly,stradi2024optimal} only for finite state-space; we obtain them for
contextual bandits with general regression oracles. Safe bandits enforce the constraint in every round with high probability and achieve zero
violation \citep{amani2019linear,pacchiano2025contextual}, but they rely on a known safe action
and confidence sets for a linear cost model, which a general regression oracle does not provide. 

\vspace{-0.05in}
\section{Problem Setup}\label{sec:setup}
\vspace{-0.03in}
\paragraph{Protocol.} Let $\mathcal{X}$ be a context space and $\mathcal{A} = [K] = \{1,\dots,K\}$ a
finite set of actions. In each round $t = 1, \dots, T$:
\vspace{-0.1in}
\begin{enumerate}
  \item nature draws a context $x_t \in \mathcal{X}$ independently from a fixed distribution
  $\mathcal{D}$ and reveals it; \vspace{-0.05in}
  \item the learner chooses a distribution $p_t(\cdot \mid x_t) \in \Delta(\mathcal{A})$ and
  samples an action $a_t \sim p_t(\cdot \mid x_t)$;\vspace{-0.05in}
  \item the learner observes a loss $\ell_t(a_t) \in [0,1]$ and a cost $c_t(a_t) \in [0,1]$ for
  the chosen action only.
  \vspace{-0.05in}
\end{enumerate}
We write $\mathcal{F}_{t-1}$ for the history up to the end of round $t-1$, that is, the contexts,
actions, losses and costs of rounds $1, \dots, t-1$.
The learner can evaluate expectations under $\mathcal{D}$; in our benchmarks, $\mathcal{D}$ is the
uniform distribution over a known finite set of start states or days. Remark~\ref{rem:unknownD}
discusses unknown $\mathcal{D}$, and Section~\ref{sec:perctx} gives a version of the algorithm
and its guarantees that does not use $\mathcal{D}$ at all and allows adversarial contexts.
\vspace{-0.1in}
\paragraph{Mean loss and cost.} The loss and cost are noisy, with conditional means
\[
  f^*_r(x,a) = \mathbb{E}[\ell_t(a) \mid x_t = x], \qquad
  f^*_c(x,a) = \mathbb{E}[c_t(a) \mid x_t = x].
\]
The learner does not know $f^*_r$ or $f^*_c$.
\vspace{-0.1in}
\paragraph{Policies and values.} A (randomized) policy $\pi$ maps each context $x$ to a
distribution $\pi(\cdot \mid x) \in \Delta(\mathcal{A})$; $p_t$ denotes the policy played in round
$t$. The expected loss and the expected cost of a policy $\pi$ are
\[
  V_r^\pi = \mathbb{E}_{x \sim \mathcal{D}}\, \mathbb{E}_{a \sim \pi(\cdot \mid x)}\big[f^*_r(x,a)\big],
  \qquad
  V_c^\pi = \mathbb{E}_{x \sim \mathcal{D}}\, \mathbb{E}_{a \sim \pi(\cdot \mid x)}\big[f^*_c(x,a)\big],
\]
both linear in $\pi$. Given a budget $b \in (0,1)$, the benchmark is the best policy that meets
the budget on average over contexts,
\[
  \pi^* \in \arg\min_{\pi} \big\{ V_r^\pi : V_c^\pi \le b \big\}.
\]
We assume the minimum is attained, which holds for instance when $\mathcal{X}$ is finite, as in
our benchmarks (the problem is then a finite linear program).
The budget is on the average cost, not on every context: $\pi^*$ may exceed $b$ at some contexts
if it stays below $b$ at others. We consider the constraint on every context in Section~\ref{sec:perctx}.
\vspace{-0.05in}
\paragraph{Performance measures.} We measure the regret and the constraint violation
\[
  \mathrm{Reg}_r(T) = \sum_{t=1}^T \big(V_r^{p_t} - V_r^{\pi^*}\big), \qquad
  \mathrm{Viol}(T) = \sum_{t=1}^T \big(V_c^{p_t} - b\big)^+ ,
\]
where $(z)^+ = \max\{z, 0\}$. The violation adds up only the rounds in which the played policy
exceeds the budget, so rounds below the budget cannot offset rounds above it. We call it
\emph{cancellation-free}, in contrast to the signed sum $\sum_t (V_c^{p_t} - b)$ that Lagrangian
methods control.
Within a round, $V_c^{p_t}$ is an average over contexts, so a policy may still exceed $b$ at some
contexts and stay below it at others; Section~\ref{sec:perctx} removes this as well.
\vspace{-0.05in}
\paragraph{Regression oracles.} The learner has access to two online regression oracles,
$\mathsf{SqAlg}_r$ for the loss and $\mathsf{SqAlg}_c$ for the cost. At the start of round $t$ they
output estimates $\hat f^t_r, \hat f^t_c : \mathcal{X}\times\mathcal{A}\to[0,1]$, computed from
$\mathcal{F}_{t-1}$, and after the round they are updated with $(x_t, a_t, \ell_t(a_t))$ and
$(x_t, a_t, c_t(a_t))$. For a policy $p$, we write
$\hat V_c^p = \mathbb{E}_{x \sim \mathcal{D}}\, \mathbb{E}_{a \sim p(\cdot \mid x)} [\hat f^t_c(x,a)]$
for its estimated cost.

We assume the followings:
\vspace{-0.1in}
\begin{assumption}[Realizability]\label{ass:real}
$f^*_r \in \mathcal{F}_r$ and $f^*_c \in \mathcal{F}_c$ for known classes
$\mathcal{F}_r, \mathcal{F}_c \subset (\mathcal{X} \times \mathcal{A} \to [0,1])$.
\end{assumption}
\vspace{-0.25in}
\begin{assumption}[Online regression oracles]\label{ass:oracle}
With probability at least $1-\zeta$,
\[
  \sum_{t=1}^T \big(\hat f^t_r - f^*_r\big)^2(x_t, a_t) \le \mathrm{Reg}^r_{\mathsf{Sq}}(T), \qquad
  \sum_{t=1}^T \big(\hat f^t_c - f^*_c\big)^2(x_t, a_t) \le \mathrm{Reg}^c_{\mathsf{Sq}}(T).
\]
For finite classes both bounds are $O(\log(|\mathcal{F}|/\zeta))$ \citep{foster2020beyond}.
\end{assumption}
We write
\[
  \mathrm{Reg}_{\mathsf{Sq}}(T)=\max\{\mathrm{Reg}^r_{\mathsf{Sq}}(T),\mathrm{Reg}^c_{\mathsf{Sq}}(T)\}+\log(1/\zeta)
\]
for the combined oracle error; it appears only in the $O(\cdot)$ forms of the bounds.
\vspace{-0.1in}
\begin{assumption}[Slater]\label{ass:slater}
There exist a policy $\pi^s$ and $\delta \in (0,1]$ with $V_c^{\pi^s} \le b - \delta$.
\end{assumption}

The Slater margin $\delta$ measures how far below the budget some policy can stay. The above assumptions are reasonable and have been used in many constrained optimization work such as \citep{ghosh2022provably,efroni2020exploration,sarkar2026simple}. We can relax this assumption to some upper bound of $\Lambda$ which ensures the admittance of a feasible solution.
\begin{lemma}[Consequences of the oracle bounds]\label{lem:oracle}
With probability at least $1-4\zeta$, for $g \in \{r, c\}$,
\begin{align*}
  \sum_{t=1}^T \mathbb{E}_{a \sim p_t(\cdot\mid x_t)}\big(\hat f^t_g - f^*_g\big)^2(x_t,a)
    &\le U_g := 2\,\mathrm{Reg}^g_{\mathsf{Sq}}(T) + 2\log(1/\zeta), \\
  \sum_{t=1}^T \big(V_c^{p_t} - \hat V_c^{p_t}\big)^2
    &\le U^{\mathcal{D}}_c := 2U_c + 2\log(1/\zeta).
\end{align*}
\end{lemma}

The detailed proof of Lemma \ref{lem:oracle} is given in Section \ref{app:setup} (Supplementary material)

\paragraph{Parameters.}
We clip the rectified weight at $\Lambda = 2/\delta$ and use the SquareCB learning rate
$\gamma = 2\sqrt{KT/(U_r + \Lambda^2 U_c)}$ \citep{foster2020beyond}.
\vspace{-0.05in}
\section{The Algorithm}
\label{sec:algo}

The algorithm is SquareCB run on the composite score
$\hat y_{t,a}=\hat f_r^t(x_t,a)+\mu_t\hat f_c^t(x_t,a)$. The only new
element is the rectified weight $\mu_t$, chosen so that the estimated cost of the played
policy meets the budget. It depends only on the current oracle estimates, so it is
available before the learner acts.
\vspace{-0.2in}
\subsection{The score-to-distribution map}
\vspace{-0.03in}
For a weight $\mu\ge0$ and a context $x$, SquareCB maps the score vector to a
distribution exactly as in Algorithm~1 of \citep{foster2020beyond}:
\vspace{-0.1in}
\begin{equation}\label{eq:p-of-mu}
  \hat y_{t,a}(\mu)=\hat f_r^t(x,a)+\mu\,\hat f_c^t(x,a),\quad
  p^{(\mu)}_{t,a}=
  \begin{cases}
    \dfrac{1}{K+\gamma(\hat y_{t,a}(\mu)-\hat y_{t,\hat a}(\mu))}, & a\ne \hat a,\\[2mm]
    1-\sum_{a'\ne \hat a}p^{(\mu)}_{t,a'}, & a=\hat a,
  \end{cases}
\end{equation}
where the leader $\hat a = \hat a_t(x,\mu)$ is a minimizer of $\hat y_{t,\cdot}(\mu)$; among tied
minimizers it is the one with the smallest $\hat f_c^t(x,\cdot)$. Write
$g_t(\mu)=\hat V_c^{p^{(\mu)}_t}=\mathbb{E}_{x\sim\mathcal{D}}\,\mathbb{E}_{a\sim p^{(\mu)}_t(\cdot\mid x)}[\hat f_c^t(x,a)]$
for the estimated cost of the resulting policy.

\vspace{-0.05in}
\begin{lemma}[Shape of $g_t$]\label{lem:shape}
$g_t$ is nonincreasing and right-continuous on $[0,\infty)$, and all its discontinuities are
downward jumps.
\end{lemma}
\vspace{-0.1in}
The complete proof has been given in Section \ref{app:algo} (Supplementary material). This will ensure that we can apply a binary search to find a suitable $\mu$ (Proposition~\ref{prop:fp}). 
\vspace{-0.1in}
\subsection{The rectified weight}
 In the definition of $p^{(\mu)}_t$ (just after \eqref{eq:p-of-mu}), we take the tied
action with the smallest estimated cost $\hat f_c^t(x,\cdot)$. We also need the distribution
that takes the tied action with the largest estimated cost: we write $\bar p^{(\mu)}_t$ for it.
Without a tie, $\bar p^{(\mu)}_t=p^{(\mu)}_t$. With a tie, as $\mu$ increases,
$\bar p^{(\mu)}_t$ is the distribution played just before the leader switches, and
$p^{(\mu)}_t$ the one played just after.
\vspace{-0.1in}

\begin{definition}[Rectified weight and played policy]\label{def:mu}
At round $t$:
\begin{enumerate}[leftmargin=*]
\item if $g_t(0)\le b$, set $\mu_t = 0$ and $p_t = p^{(0)}_t$;\vspace{-0.05in}
\item if $g_t(\Lambda)>b$, set $\mu_t = \Lambda$ and $p_t = p^{(\Lambda)}_t$;\vspace{-0.05in}
\item otherwise, let $\mu_t = \inf\{\mu\in[0,\Lambda] : g_t(\mu)\le b\}$ and play
  $p_t = w_t\,\bar p^{(\mu_t)}_t + (1-w_t)\,p^{(\mu_t)}_t$, with $w_t\in[0,1]$ chosen so that
  $\hat V_c^{p_t} = b$.\vspace{-0.08in}
\end{enumerate}
\end{definition}

In the third case, the mixture is needed because $g_t$ can jump over $b$ at $\mu_t$
(Lemma~\ref{lem:shape}): then $p^{(\mu_t)}_t$ has estimated cost strictly below $b$,
$\bar p^{(\mu_t)}_t$ has estimated cost above $b$, and no single weight gives exactly $b$.
Mixing the two gives exactly $b$, which the analysis uses (Lemma~\ref{lem:extract}). If $g_t$
is continuous at $\mu_t$, the two distributions have the same estimated cost $b$ and the
mixture plays no role.
\vspace{-0.05in}
\begin{proposition}[Existence, computability]\label{prop:fp}
The policy $p_t$ of Definition~\ref{def:mu} is well defined, and in the third case
$\hat V_c^{p_t} = b$. It can be computed by bisection on $[0,\Lambda]$. In every case $p_t$ is,
for each $x$, a convex combination of inverse-gap-weighted distributions for the score
vector $\hat y_t(\mu_t)$, each with a minimizer of $\hat y_t(\mu_t)$ as the leader. Finally,
$\mu_t$ and $p_t$ depend only on $\hat f_r^t, \hat f_c^t$, hence are
$\mathcal F_{t-1}$-measurable; in particular they do not depend on the realized action $a_t$.
\end{proposition}
\vspace{-0.11in}
Proof for Proposition \ref{prop:fp} is given in Section \ref{app:prop_fp} (Supplementary material)
\vspace{-0.1in}
\begin{center}
\fbox{\parbox{0.93\textwidth}{
\textbf{Algorithm: Cancellation-Free Constrained SquareCB.}\\
Parameters $\Lambda=2/\delta$, $\gamma=2\sqrt{KT/(U_r+\Lambda^2U_c)}$, budget $b$; oracles
$\mathsf{SqAlg}_r,\mathsf{SqAlg}_c$.\\[3pt]
For $t=1,\dots,T$:
\begin{enumerate}
  \item Receive $x_t$.\vspace{-0.05in}
  \item Compute $\mu_t$ and $p_t$ (Definition~\ref{def:mu}) by bisection on $g_t(\cdot)$,
        mixing the two policies at the end of the bisection.\vspace{-0.05in}
  \item Sample $a_t\sim p_t(\cdot\mid x_t)$; observe $\ell_t(a_t),c_t(a_t)$;
        update both oracles.\vspace{-0.05in}
\end{enumerate}
}}
\end{center}
\vspace{-0.1in}
\section{The SquareCB Guarantee}
\label{sec:scb}
\vspace{-0.03in}
\begin{lemma}[IGW inequality \citep{foster2020beyond}]\label{lem:igw}
Let $p$ be inverse-gap weighting for a score vector $\hat y\in\mathbb{R}^K$, with any minimizer
of $\hat y$ as the leader. Then for every $f\in\mathbb{R}^K$,
\begin{equation}\label{eq:lem3}
  \sum_a p_a f_a - \min_a f_a \le \frac{2K}{\gamma} + \frac{\gamma}{4}\sum_a p_a(\hat y_a - f_a)^2 .
\end{equation}
\end{lemma}
\citet[Lemma~3]{foster2020beyond} state it for vectors in $[0,1]^K$; the proof uses no range
assumption (Appendix~\ref{app:scb}), so it applies to our composite scores in
$[0,1+\Lambda]^K$. Both sides are linear in $p$ for fixed $(\hat y,f)$, so it also holds for
the mixtures of Definition~\ref{def:mu}.

\begin{lemma}[Composite regret]\label{lem:sqcb}
For any fixed comparator policy $\pi$, with probability at least $1-5\zeta$,
\begin{equation}\label{eq:sqcb1}
  \sum_{t=1}^T\Bigl[(V_r^{p_t}+\mu_t V_c^{p_t})-(V_r^{\pi}+\mu_t V_c^{\pi})\Bigr]
  \le R_T:=2\sqrt{KT\,(U_r+\Lambda^2U_c)} + (1+\Lambda)\sqrt{2T\log(1/\zeta)} .
\end{equation}
With $\Lambda=2/\delta$, $R_T=O\bigl(\sqrt{KT\,\mathrm{Reg}_{\mathsf{Sq}}(T)}/\delta\bigr)$.
\end{lemma}

The proof is given in Section \ref{app:scb} in Supplimentary file.

\vspace{-0.1in}
\section{Master Inequality and the Two Bounds}
\label{sec:mi2b}
Before stating and proving the main results, we state two important results. 
Throughout write $E_c:=\sqrt{T\,U^{\mathcal{D}}_c}$; by Cauchy--Schwarz and
Lemma~\ref{lem:oracle}, $\sum_t|V_c^{p_t}-\hat V_c^{p_t}|\le E_c$.

\begin{lemma}[Rectified extraction]\label{lem:extract}
With $\mu_t$ and $p_t$ from Definition~\ref{def:mu},
\begin{equation}\label{eq:extract1}
  \sum_{t=1}^T\mu_t\bigl(V_c^{p_t}-b\bigr)\;\ge\;\Lambda\,\mathrm{Viol}(T)-2\Lambda E_c .
\end{equation}
\end{lemma}
The proof is given in Section \ref{app:mi2b} (in Supplementary material)





\begin{remark}[Approximate matching]\label{rem:approx}
The third case of Definition~\ref{def:mu} asks for $\hat V_c^{p_t}=b$ exactly. Suppose instead
that only $|\hat V_c^{p_t}-b|\le\eta_t$ holds there, for example because of a bisection
tolerance. If $\hat V_c^{p_t}\le b$, then $V_c^{p_t}-b\le\epsilon_t$ and the left side is at least
$-\Lambda(\epsilon_t+\eta_t)$; if $\hat V_c^{p_t}\ge b$, then $V_c^{p_t}-b\ge-\epsilon_t$ and
$(V_c^{p_t}-b)^+\le\epsilon_t+\eta_t$. In both cases the per-round inequality in the proof of
Lemma~\ref{lem:extract} holds with an extra $-\Lambda\eta_t$, so \eqref{eq:extract1} holds with
$-\Lambda\sum_t\eta_t$ added to its right side. Following this through, the violation bound
of Theorem~\ref{thm:viol} increases by $2\sum_t\eta_t$ and the regret bound of
Theorem~\ref{thm:reward} by $\Lambda\sum_t\eta_t$, so any tolerance with
$\sum_t\eta_t=O(\sqrt T)$, such as $\eta_t=t^{-1/2}$, keeps both rates. 
\end{remark}

\begin{lemma}[Master inequality]\label{lem:master}
On the event of Lemma~\ref{lem:sqcb} with $\pi = \pi^*$,
\begin{equation}\label{eq:master}
  \sum_{t=1}^T\bigl(V_r^{p_t}-V_r^{\pi^*}\bigr)+\Lambda\,\mathrm{Viol}(T)
  \;\le\; R_T+2\Lambda E_c .
\end{equation}
\end{lemma}

\begin{proof}
Apply \eqref{eq:sqcb1} with comparator $\pi=\pi^*$. Since
$V_c^{\pi^*}\le b$ and $\mu_t\ge0$,
$\mu_t(V_c^{p_t}-V_c^{\pi^*})\ge\mu_t(V_c^{p_t}-b)$, so
\begin{equation}\label{eq:starstar}
  \sum_t(V_r^{p_t}-V_r^{\pi^*})+\sum_t\mu_t(V_c^{p_t}-b)\le R_T .
\end{equation}
Substituting the extraction \eqref{eq:extract1} in Lemma~\ref{lem:extract} gives \eqref{eq:master}.
\end{proof}
Now, we are ready to state and prove the main results:
\begin{theorem}[Cancellation-free violation]\label{thm:viol}
Under Assumptions~\ref{ass:real}--\ref{ass:slater}, with $\Lambda=2/\delta$, with probability
at least $1-5\zeta$,
\[
  \mathrm{Viol}(T)\le \delta R_T + 4E_c = O\!\left(\sqrt{KT\,\mathrm{Reg}_{\mathsf{Sq}}(T)}\right).
\]
\end{theorem}

\begin{proof}
By Lemma~\ref{lem:qsaddle},
$\sum_t(V_r^{p_t}-V_r^{\pi^*})\ge-\lambda^*\mathrm{Viol}(T)$. Substituting into \eqref{eq:master} in Lemma~\ref{lem:master},
$(\Lambda-\lambda^*)\,\mathrm{Viol}(T)\le R_T+2\Lambda E_c$.
Since $\lambda^*\le1/\delta$ and $\Lambda=2/\delta$, $\Lambda-\lambda^*\ge1/\delta$, so
$\mathrm{Viol}(T)\le\delta R_T+2\Lambda\delta E_c=\delta R_T+4E_c$. Finally
$\delta R_T \le 2\sqrt{KT(\delta^2U_r+4U_c)} + (2+\delta)\sqrt{2T\log(1/\zeta)}$ and
$E_c = O(\sqrt{T\,\mathrm{Reg}_{\mathsf{Sq}}})$.
\end{proof}
\vspace{-0.1in}
\begin{theorem}[Reward regret]\label{thm:reward}
Under the same assumptions, with probability at least $1-5\zeta$,
\[
  \mathrm{Reg}_r(T)\le R_T + 2\Lambda E_c = O\!\left(\frac{\sqrt{KT\,\mathrm{Reg}_{\mathsf{Sq}}(T)}}{\delta}\right).
\]
\end{theorem}

\begin{proof}
From \eqref{eq:starstar},
$\mathrm{Reg}_r(T)\le R_T-\sum_t\mu_t(V_c^{p_t}-b)$, and by
\eqref{eq:extract1}, $\sum_t\mu_t(V_c^{p_t}-b)\ge\Lambda\mathrm{Viol}(T)-2\Lambda E_c\ge-2\Lambda E_c$.
\end{proof}
\vspace{-0.1in}
\begin{corollary}[Cancellation-free regret]\label{cor:posreg}
On the event of Theorems~\ref{thm:viol} and~\ref{thm:reward},
\[
  \sum_{t=1}^T\bigl(V_r^{p_t}-V_r^{\pi^*}\bigr)^+\le\mathrm{Reg}_r(T)+\frac{\mathrm{Viol}(T)}{\delta}
  =O\!\left(\frac{\sqrt{KT\,\mathrm{Reg}_{\mathsf{Sq}}(T)}}{\delta}\right).
\]
\end{corollary}

\begin{proof}
Write $u_t=V_r^{p_t}-V_r^{\pi^*}$, so $u_t^+=u_t+(-u_t)^+$. By Lemma~\ref{lem:qsaddle},
$-u_t\le\lambda^*(V_c^{p_t}-b)^+$, and the right side is nonnegative, so
$(-u_t)^+\le\lambda^*(V_c^{p_t}-b)^+$. Summing and using $\lambda^*\le1/\delta$ gives
$\sum_tu_t^+\le\mathrm{Reg}_r(T)+\mathrm{Viol}(T)/\delta$; Theorems~\ref{thm:viol}
and~\ref{thm:reward} give the rate.
\end{proof}

In words, the played policy can beat $\pi^*$ only in rounds where it exceeds the budget, and by
at most $\lambda^*$ times the excess, so negative-regret rounds cannot hide more than
$\mathrm{Viol}(T)/\delta$; the same argument with $\lambda^*(x_t)$ applies per context
(Section~\ref{sec:perctx}). The $1/\delta$ factor in the regret is common in constrained
learning, where Slater's condition bounds the optimal multiplier
\citep{slivkins2023contextual,ghosh2022provably}; the violation bound does not
depend on $\delta$. For comparison, \citep{sarkar2026simple} obtain regret and signed violation
$O(\sqrt{K}T^{3/4}U_T^{1/4})$ against the same feasible-in-expectation benchmark with adversarial
contexts, and $\sqrt{T}$ regret only for the deterministic action space (no randomization over the actions are permitted) \citep[Theorem~9]{sarkar2026simple}.



\begin{remark}[Unknown context distribution]\label{rem:unknownD}
If $\mathcal{D}$ is unknown, $g_t$ can be computed with the empirical distribution of
$x_1,\dots,x_{t-1}$. The analysis then needs an extra term for the difference between
$\hat V_c^{p_t}$ under the empirical and the true distribution, uniformly over the policies the
algorithm can play; we do not pursue this here.
\end{remark}
\vspace{-0.2in}
\section{Per-Context Constraints}\label{sec:perctx}
\vspace{-0.03in}
We extend our proofs to per context guarantees. Let the contexts $x_1, x_2, \dots$ be arbitrary, possibly chosen
by an adaptive adversary, and replace the benchmark and measures by
\[
  \mathrm{OPT}(x) = \min_{p\in\Delta(\mathcal{A})}\bigl\{\langle p, f^*_r(x,\cdot)\rangle :
    \langle p, f^*_c(x,\cdot)\rangle \le b\bigr\},
\]
\[
  \mathrm{Reg}^{\mathrm{ctx}}(T)=\sum_{t=1}^T\bigl(\langle p_t, f^*_r(x_t,\cdot)\rangle - \mathrm{OPT}(x_t)\bigr),\qquad
  \mathrm{Viol}^{\mathrm{ctx}}(T)=\sum_{t=1}^T\bigl(\langle p_t, f^*_c(x_t,\cdot)\rangle - b\bigr)^+,
\]
where $p_t = p_t(\cdot\mid x_t)$. Assume per-context Slater: for every $x$ some
$p^s(x)$ has $\langle p^s(x), f^*_c(x,\cdot)\rangle \le b-\delta$. The algorithm is unchanged
except that $g_t(\mu) = \langle p^{(\mu)}_t(\cdot\mid x_t), \hat f^t_c(x_t,\cdot)\rangle$ specific to one context

\begin{theorem}[Per-context guarantees]\label{thm:ctx}
Under Assumptions~\ref{ass:real}--\ref{ass:oracle} and per-context Slater, with
$\Lambda = 2/\delta$, with probability at least $1-3\zeta$,
\[
  \mathrm{Reg}^{\mathrm{ctx}}(T) \le R'_T + 2\Lambda E'_c, \qquad
  \mathrm{Viol}^{\mathrm{ctx}}(T) \le \delta R'_T + 4E'_c,
\]
with $R'_T = 2\sqrt{KT(U_r+\Lambda^2U_c)}$ and $E'_c = \sqrt{T U_c}$; that is,
$\mathrm{Reg}^{\mathrm{ctx}}(T) = O(\sqrt{KT\,\mathrm{Reg}_{\mathsf{Sq}}}/\delta)$ and
$\mathrm{Viol}^{\mathrm{ctx}}(T) = O(\sqrt{KT\,\mathrm{Reg}_{\mathsf{Sq}}})$.
\end{theorem}
\vspace{-0.05in}
\begin{proof}
Repeat the proofs above with $V^{p}_g$ replaced by $\langle p(\cdot\mid x_t), f^*_g(x_t,\cdot)\rangle$,
$\pi^*$ by the per-context optimum and $\lambda^*$ by the per-context multiplier
$\lambda^*(x_t)\le1/\delta$. Lemma~\ref{lem:sqcb} is applied at $x_t$ and needs no
conversion to an expectation over $\mathcal{D}$, so the Azuma term disappears. The
estimation error is
$|\langle p_t, \hat f^t_c - f^*_c\rangle(x_t)| \le (\mathbb{E}_{a\sim p_t}(\hat f^t_c - f^*_c)^2(x_t,a))^{1/2}$
by Jensen, so $\sum_t\epsilon_t \le \sqrt{TU_c}$ by Cauchy--Schwarz and the first bound of
Lemma~\ref{lem:oracle}. The comparator step of Lemma~\ref{lem:master} uses
$\langle \pi^*(x_t), f^*_c(x_t,\cdot)\rangle \le b$, which holds because the benchmark is
per-context.
\end{proof}

 The average version needs only average Slater and
compares with a stronger benchmark, but needs i.i.d.\ contexts and access to $\mathcal{D}$,
and lets the policy exceed $b$ in some contexts. The per-context version needs a policy that is
safe in every context, which in control is usually a known safe fallback controller, and in
return holds for arbitrary context sequences and bounds the violation in every episode.
\vspace{-0.2in}
\section{Experiments}
\vspace{-0.03in}
We evaluate our method on two control benchmarks, the 2D quadrotor of safe-control-gym and the
BOPTEST building, and compare it with primal-dual and with the CCB reduction scheme of
\citep{sarkar2026simple}. \emph{Note although \citep{deb2025conservative} has some notion of safety and uses IGW over an online regression oracle, the formulation is not explicitly modelled to work for constrained bandit setup. Further theoretical guarantees also differs from our objective in this work. Hence, we do not compare with that setup. We rather compare with \citep{sarkar2026simple} which is more aligned to our objective.} In both benchmarks, each round is one episode: the learner chooses a controller at the start of the episode, the controller runs for the whole episode, and the loss and the constraint cost are observed at its end.
\vspace{-0.05in}
\paragraph{Quadrotor.} We use the 2D quadrotor of safe-control-gym \citep{yuan2022safe}.
In each episode, the learner observes the drone's start state and payload mass, drawn from
600 start states, and chooses one of six LQR controllers to bring it to a height of 1\,m,
while random disturbance forces act on it. Fast controllers track well but overshoot the
target, and slow ones overshoot little but track worse. The loss is the tracking error, the
cost is the overshoot (how far the drone rises above the target), and the budget is an average
overshoot of 1.75\,cm.

\vspace{-0.05in}
\paragraph{Building.} We use the \texttt{bestest\_air} test case of BOPTEST \citep{blum2021building},
a single zone with a fan-coil unit. Each day, drawn from 180 days of the year, the learner
observes the weather forecast and chooses one of eight thermostat schedules. The building's
thermal state at the start of the day is random and not observed. Energy-saving schedules use
less heating and cooling but let the temperature leave the comfort band, while comfortable
schedules use more energy. The loss is the energy use, the cost is the thermal discomfort
(Kelvin-hours outside 21--25.5\,$^\circ$C), and the budget is 10\,K$\cdot$h per day.
\vspace{-0.2in}
\section{Results}\label{sec:results}
\begin{figure}[htbp]
\vspace{-0.1in}
    \centering
    \begin{minipage}{0.4\textwidth}
        \centering
        \includegraphics[width=\linewidth]{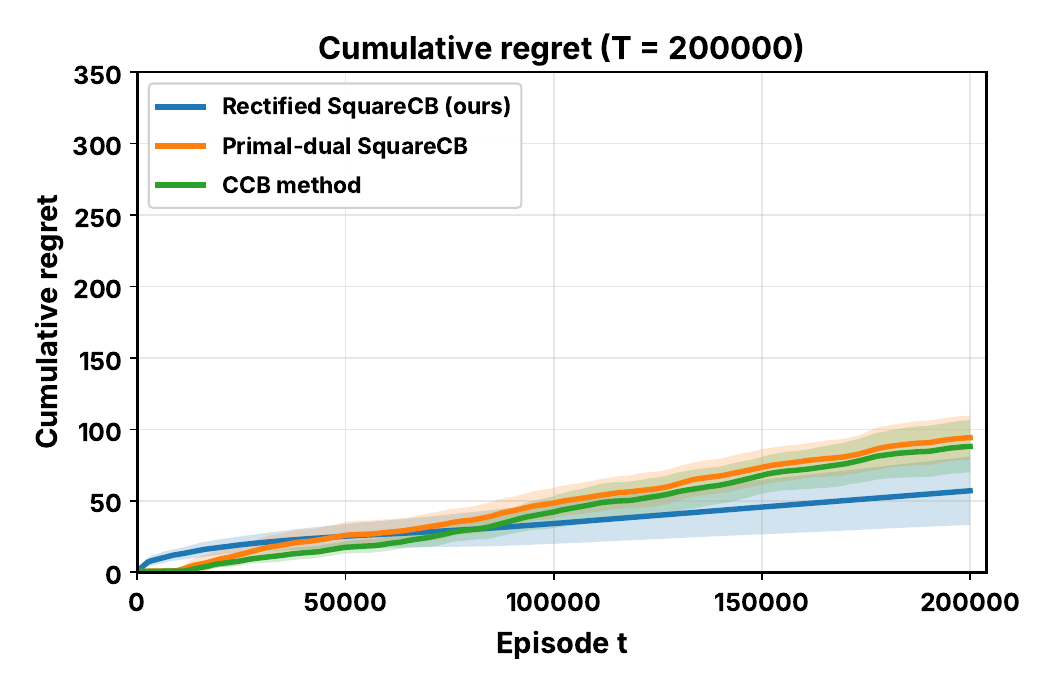}
    \end{minipage}
    \hfill
    \begin{minipage}{0.4\textwidth}
        \centering
        \includegraphics[width=\linewidth]{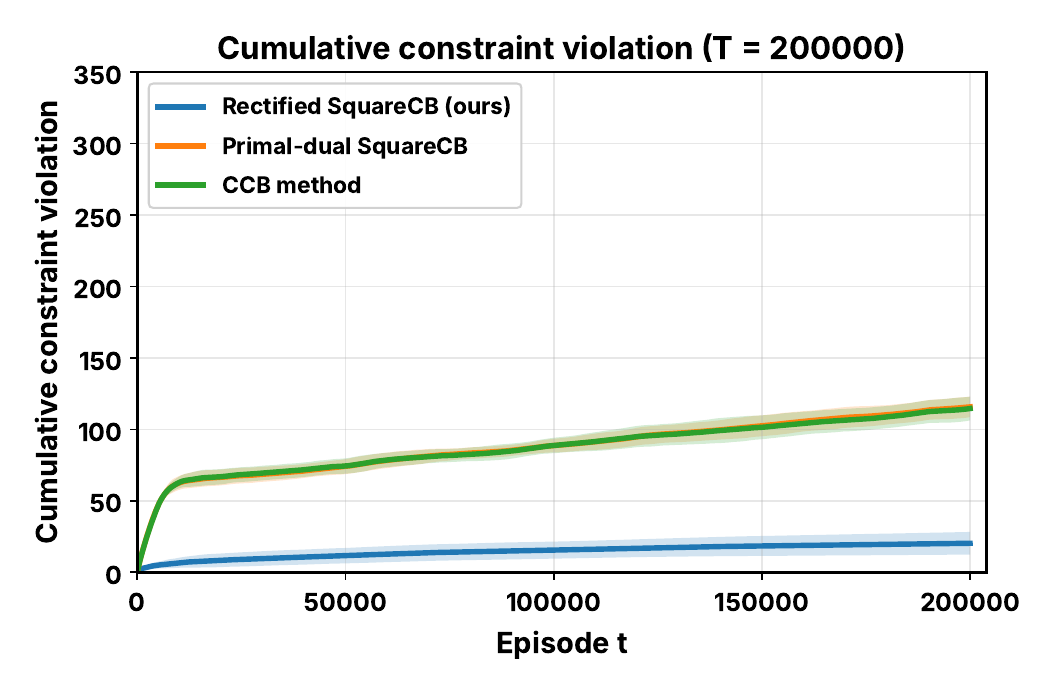}
    \end{minipage}
    \caption{\footnotesize Quadrotor:cumulative regret and violation (both cancellation-free) averaged over 10 runs.}
    \label{fig:quad}
\end{figure}

\begin{figure}[htbp]
    \centering
\vspace{-0.1in}
    \begin{minipage}{0.4\textwidth}
        \centering
        \includegraphics[width=\linewidth]{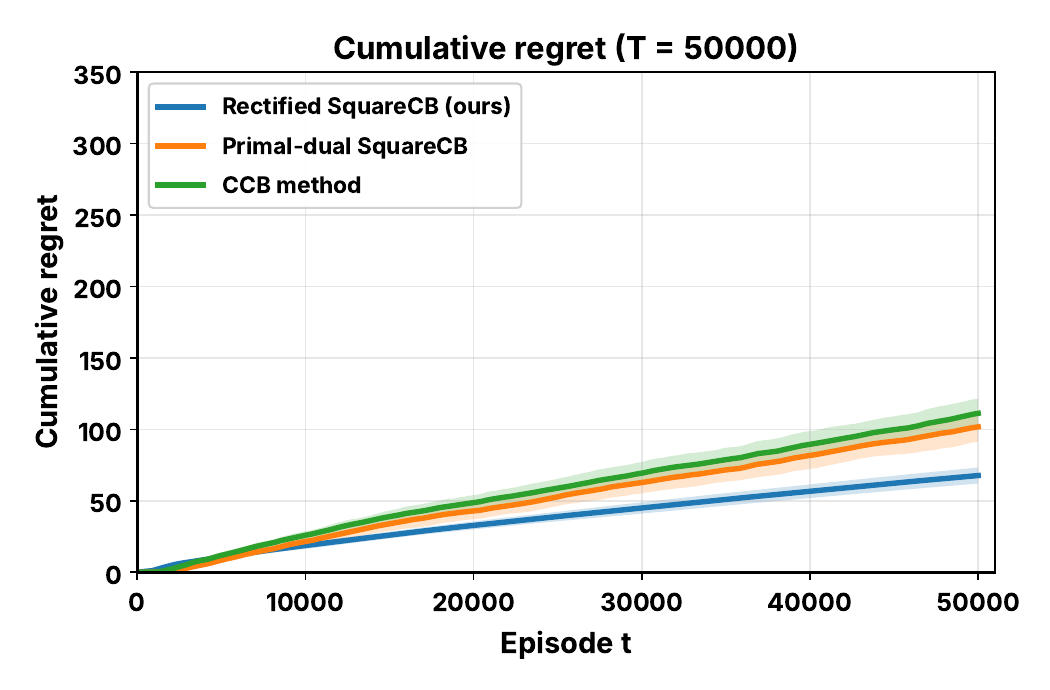}
    \end{minipage}
    \hfill
    \begin{minipage}{0.4\textwidth}
        \centering
        \includegraphics[width=\linewidth]{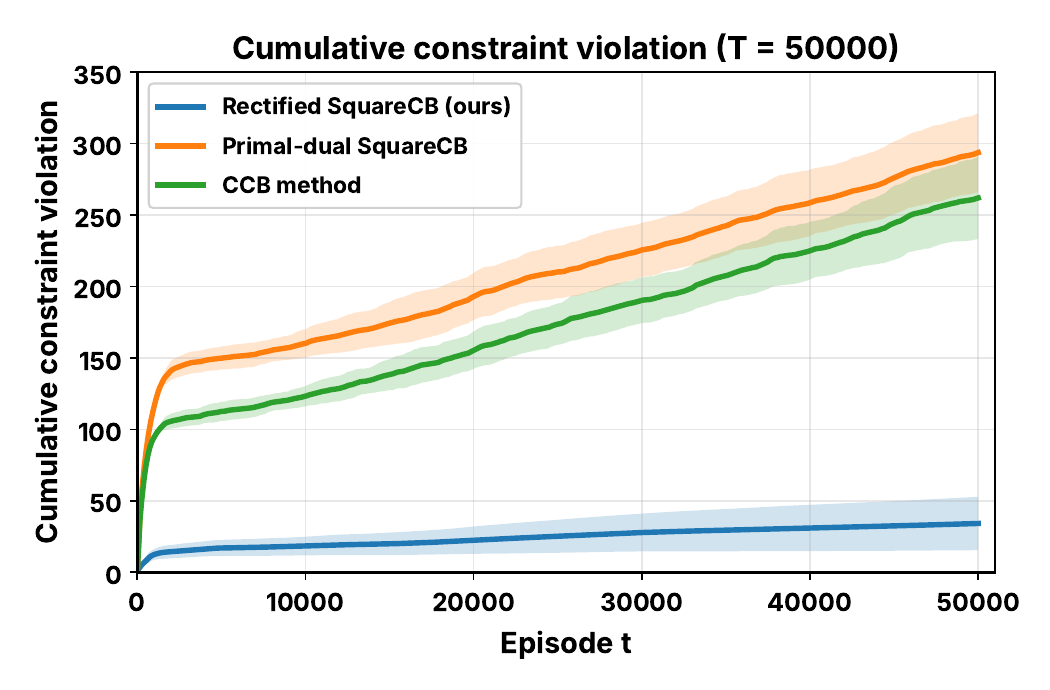}
    \end{minipage}
\vspace{-0.1in}
    \caption{\footnotesize Building:cumulative regret and violation (left and right, respectively) averaged over 10 runs.}
\vspace{-0.15in}
    \label{fig:building}
\end{figure}

\vspace{-0.03in}
Figures~\ref{fig:quad} and~\ref{fig:building} show that for each benchmark, the
cumulative regret (left) and the cumulative constraint violation (right) as functions of the
number of episodes. Each curve is the average over ten runs, and the shaded band shows
$\pm 2$ standard errors. For regret we plot
$\sum_t (V_r^{p_t} - V_r^{\pi^*})^+$ (the cancellation-free bound) and the violation is also cancellation-free
$\mathrm{Viol}(T)$.
\vspace{-0.1in}
\paragraph{Quadrotor.} On the quadrotor (Figure~\ref{fig:quad}), the left plot shows the regret. The baselines
have lower regret at the start, while they exceed the budget, but their regret keeps growing, and at the final episode our method has the lowest regret (57, against 94 for primal-dual and 88 for the CCB scheme). The right plot of Figure~\ref{fig:quad} shows the violation. Both baselines have significantly larger violations compared to our method. Our method achieves sub-linear regret and violation bound  confirming the theoretical insights, and achieve a better regret without sacrificing feasibility.
\vspace{-0.07in}
\paragraph{Building.} On the building, the three methods have similar regret at the start (left of Figure~\ref{fig:building}), and from about 7{,}000 episodes on our method has the lowest regret showing that our approach can track the optimal policy better compared to the baselines similar to the Quadrotor. For the violation (right of Figure~\ref{fig:quad}), the difference is more pronounced and ours achieve significantly lower violation indicating that we achieve a better regret without sacrificing feasibility.

\vspace{-0.1in}
\paragraph{Why our method does better.} The baselines set their cost weight from accumulated
signed residuals, $\mu_{t+1}=\max(\mu_t+\eta(\hat V_c^{p_t}-b),0)$ and $w_t=2Q_{t-1}/V$ with
$Q_{t-1}=\sum_{s<t}(c_s-b)$, whereas our weight depends only on the current estimates. Their
weight starts at zero, so they first exceed the budget, which keeps their early regret low (a
policy can beat $\pi^*$ only by exceeding the budget, Lemma~\ref{lem:qsaddle}). On the
quadrotor, it takes about 10{,}000 episodes to settle, a phase that accounts for more than half
of their violation; ours settles within 1{,}000. Afterwards their weight swings with about three
times the spread of ours, and their policy crosses the budget about 4 times (quadrotor) and 9
times (building) as often: swings up make the policy too conservative and add to the regret,
swings down add to the violation. Our mechanism tries to find the policy with the right balance from the first episode onwards,  therefore gives both a lower violation
and a lower regret.

\vspace{-0.15in}
\section{Conclusion}
\vspace{-0.03in}
We studied constrained contextual bandits with general online regression oracles and proposed
a constrained version of SquareCB whose cost weight is rectified: it is zero when the estimated
cost is within the budget, and otherwise just large enough to bring it back to the budget.
Because the weight depends only on the current estimates, slack from past rounds is never
carried over, and the algorithm attains regret $O(\sqrt{KT\,\mathrm{Reg}_{\mathsf{Sq}}}/\delta)$
and cancellation-free violation $O(\sqrt{KT\,\mathrm{Reg}_{\mathsf{Sq}}})$, both for budgets on
average over contexts and, with adversarial contexts, for budgets in every context. On
controller selection for a quadrotor and a building, it exceeds the budget far less than
primal-dual and virtual-queue methods, with lower regret. The algorithm requires a known bound
on the optimal multiplier, which a safe fallback controller provides through its margin
$\delta$; removing this requirement without delaying enforcement of the budget, and extending
the method to multiple constraints, are directions for future work. Extending the approach to the Reinforcement Learning setting and more long-horizon control tasks is also left for the future.

\clearpage
\subsection*{Generative AI Use}
LLMs were used for spel-check and improving the quality of writing. The authors reviewed and verified all AI-assisted content and take responsibility for the final content of the paper.
\acks
{This research was partially supported by the National Science Foundation (NSF EPCL) Award Nos.~2534262 and 2621763.}

\bibliography{l4dc2027-sample}

\clearpage
\appendix
\begin{center}
    {\Large\bfseries Appendix}
\end{center}

\medskip
\tableofcontents


\begin{table}[t]
\floatconts
  {tab:summary}
  {\caption{Summary of the results.}}
  {\renewcommand{\arraystretch}{1.3}%
\begin{tabular}{ll}
\toprule
Result & Statement\\
\midrule
Lem.~\ref{lem:oracle} & oracle bounds $\Rightarrow$ policy-value error $\sum_t(V_c^{p_t}-\hat V_c^{p_t})^2\le U^{\mathcal D}_c$\\
Lem.~\ref{lem:shape}, Prop.~\ref{prop:fp} & $g_t$ monotone with downward jumps; rectified weight exists, computable\\
Lem.~\ref{lem:sqcb} & composite SquareCB: $\le R_T=O(\Lambda\sqrt{KT\,\mathrm{Reg}_{\mathsf{Sq}}})$\\
Lem.~\ref{lem:qsaddle} & saddle: $V_r^\pi-V_r^{\pi^*}\ge-\lambda^*(V_c^\pi-b)^+$\\
Lem.~\ref{lem:extract} & $\sum_t\mu_t(V_c^{p_t}-b)\ge\Lambda\mathrm{Viol}(T)-2\Lambda E_c$\\
Lem.~\ref{lem:master} & $\sum_t(V_r^{p_t}-V_r^{\pi^*})+\Lambda\mathrm{Viol}(T)\le R_T+2\Lambda E_c$\\
Thm.~\ref{thm:viol} & $\mathrm{Viol}(T)=O(\sqrt{KT\,\mathrm{Reg}_{\mathsf{Sq}}})$, cancellation-free\\
Thm.~\ref{thm:reward} & $\mathrm{Reg}_r(T)=O(\sqrt{KT\,\mathrm{Reg}_{\mathsf{Sq}}}/\delta)$\\
Thm.~\ref{thm:ctx} & same rates per context, for adversarial contexts\\
\bottomrule
\end{tabular}}
\end{table}


\section{Summary of Results}
Summary of Results is in Table~\ref{tab:summary}

\section{Effect of the Budget $b$}\label{app:budget}
To study how our method responds to the budget, we run it with its hyperparameters fixed at
the tuned values and vary only the budget $b$ 
(\figureref{fig:budget-quad,fig:budget-bld}).

\paragraph{In Quadrotor:}the range of meaningful budgets is set by the design of the task, namely the set of controllers and the distribution
of initial states and payloads. Choosing, in every context, the controller with the smallest expected overshoot gives
$V_c^{\min} \approx 0.44$\,cm, so Assumption~\ref{ass:slater} holds for every $b > 0.44$\,cm; the loss-minimising
policy, which ignores the constraint, overshoots by about $2.0$\,cm on average, so the constraint is active for every
$b < 2.0$\,cm. The main budget $b = 1.75$\,cm lies inside this range. To check that our conclusions do not hinge on
this choice, we run our method at $b \in \{0.5, 1, 1.5, 2\}$\,cm, from nearly infeasible to nearly inactive. The
exploration constant $\gamma_0$ keeps the value tuned for $b = 1.75$\,cm, and only the clip level $\Lambda = 2/\delta$
changes with $b$, as the method prescribes ($T = 2\cdot 10^4$, five runs).

\begin{figure}[p]
\floatconts
  {fig:budget-quad}
  {\caption{Quadrotor: effect of the budget $b$ on our method. Left: running average of the objective loss (dashed: loss of $\pi^*(b)$).
   Right: running average of the constraint cost (dashed: budget $b$). Mean $\pm$ 2 standard
   errors over 5 runs.}}
  {%
    \subfigure[Rectified SquareCB (ours)]{\label{fig:bq-ours}%
      \includegraphics[width=1.05\linewidth]{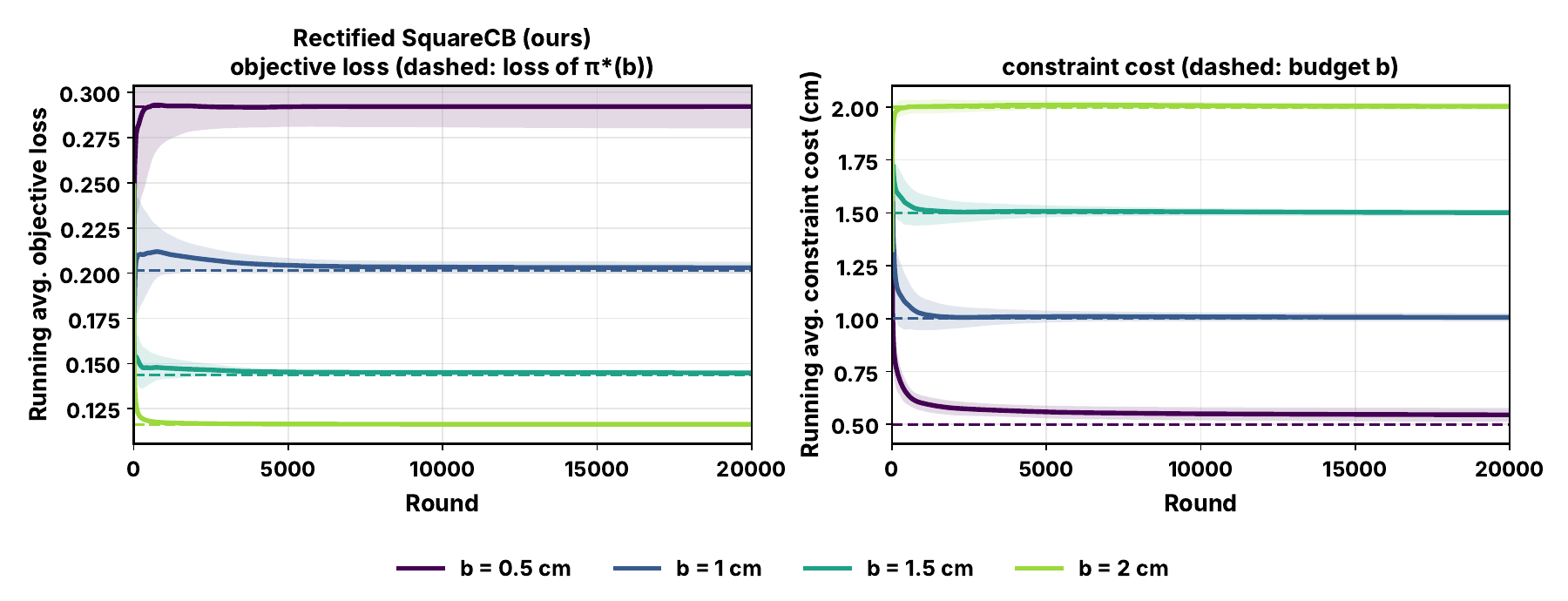}}\\[1mm]
  }
\end{figure}

Figure~\ref{fig:budget-quad} shows that for $b \in \{1, 1.5, 2\}$\,cm the average overshoot settles at the budget
within a few thousand rounds and the loss converges to that of $\pi^\star(b)$, without any retuning, in line with
Theorems~\ref{thm:viol} and~\ref{thm:reward}. Tightening the budget raises the optimal loss, from $0.116$ at
$b = 2$\,cm to $0.292$ at $b = 0.5$\,cm, and the multiplier $\lambda^\star$ with it, from $0.11$ to $6.04$; at every
budget $\lambda^\star$ stays well below $1/\delta$, as Lemma~\ref{lem:lamstar} predicts, so the default
$\Lambda = 2/\delta$ is always large enough without knowledge of $\lambda^\star$. The regret at $T = 2\cdot10^4$ grows
as the budget tightens, from $2$ at $b = 2$\,cm to $104$ at $b = 0.5$\,cm, together with $\Lambda$, consistent with
the reward bound of Theorem~\ref{thm:reward}, which grows linearly in $\Lambda$. At the tightest budget, $b = 0.5$\,cm, where $\delta$ is smallest ($\Lambda = 2/\delta \approx 700$), convergence is slower: at $T = 2\cdot10^4$ the average overshoot is within $0.05$\,cm of $b$.

\paragraph{In Building:} the range of meaningful budgets is set by the design of the task. Choosing, in every context,
the controller with the smallest expected discomfort gives $V_c^{\min} \approx 0.01$\,K$\cdot$h/day, so
Assumption~\ref{ass:slater} holds for essentially every $b > 0$; the loss-minimising policy, which ignores the
constraint, incurs about $34.9$\,K$\cdot$h/day, so the constraint is active for every $b < 34.9$\,K$\cdot$h/day. Around
the main budget $b = 10$\,K$\cdot$h/day we run our method at $b \in \{2, 20, 30\}$\,K$\cdot$h/day, from tight to
loose, with $\gamma_0$ kept at its tuned value and only $\Lambda = 2/\delta$ changing with $b$ ($T = 2\cdot10^4$,
five runs).

\begin{figure}[p]
\floatconts
  {fig:budget-bld}
  {\caption{Building: effect of the budget $b$ on our method. Left: running average of the objective loss (dashed: loss of $\pi^*(b)$).
   Right: running average of the constraint cost (dashed: budget $b$). Mean $\pm$ 2 standard
   errors over 5 runs.}}
  {%
    \subfigure[Rectified SquareCB (ours)]{\label{fig:bb-ours}%
      \includegraphics[width=1.05\linewidth]{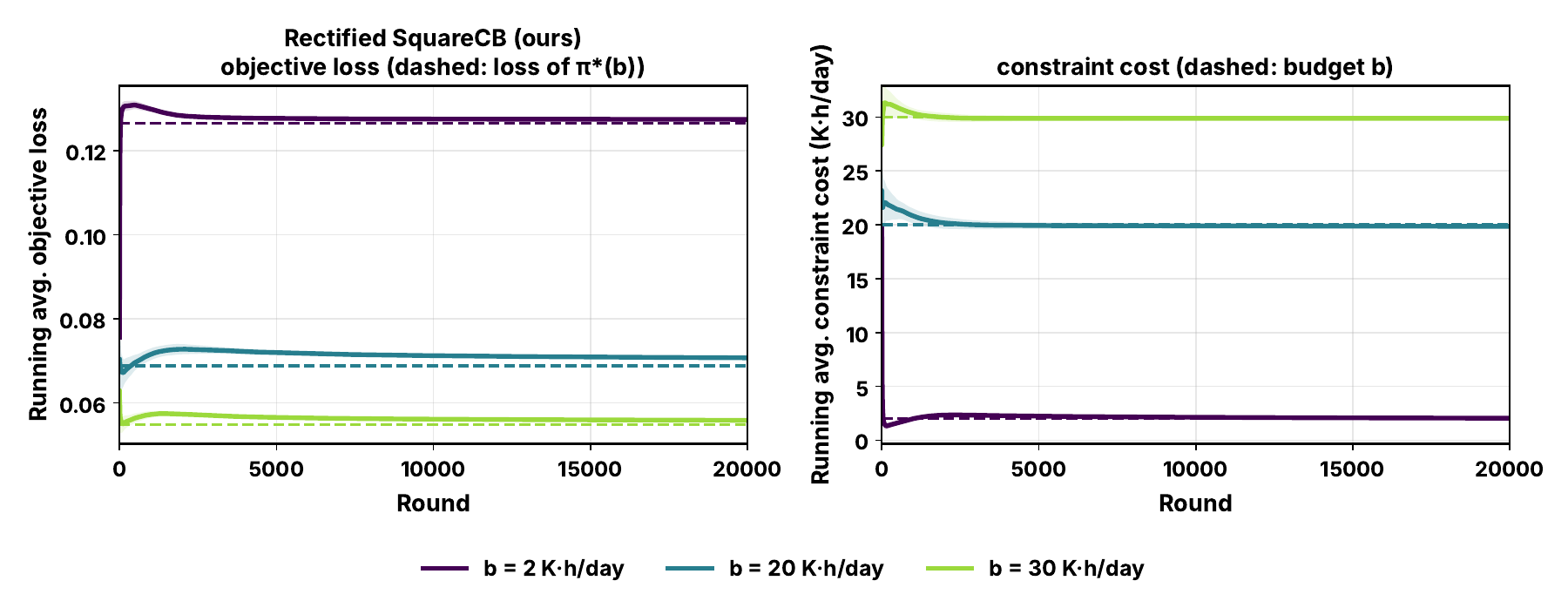}}\\[1mm]
  }
\end{figure}

Figure~\ref{fig:budget-bld} shows that the average discomfort first exceeds the budget while the costs of the controllers are being learned and then settles at $b$, without any retuning: within about $600$ and $1500$ rounds for $b = 30$ and $20$\,K$\cdot$h/day, and more slowly for the tight budget $b = 2$\,K$\cdot$h/day, where the early excess is large relative to $b$. At $T = 2\cdot10^4$ all three runs are within $2\%$ of their budget, in line with Theorem~\ref{thm:viol}. The loss stays within $3\%$ of that of $\pi^\star(b)$ and the gap is still shrinking, in line with Theorem~\ref{thm:reward}. Tightening the budget raises the optimal loss, from $0.055$ at $b = 30$ to $0.127$ at $b = 2$\,K$\cdot$h/day, and $\lambda^\star$ with it, from $0.10$ to $0.34$; at every budget $\lambda^\star$ stays well
below $1/\delta$, as Lemma~\ref{lem:lamstar} predicts, so the default $\Lambda = 2/\delta$ is large enough without knowledge of $\lambda^\star$.

\section{Ablation study on 
$\Lambda$}\label{sec:lambda}

\begin{figure}[t]
  \centering
  \includegraphics[width=\linewidth]{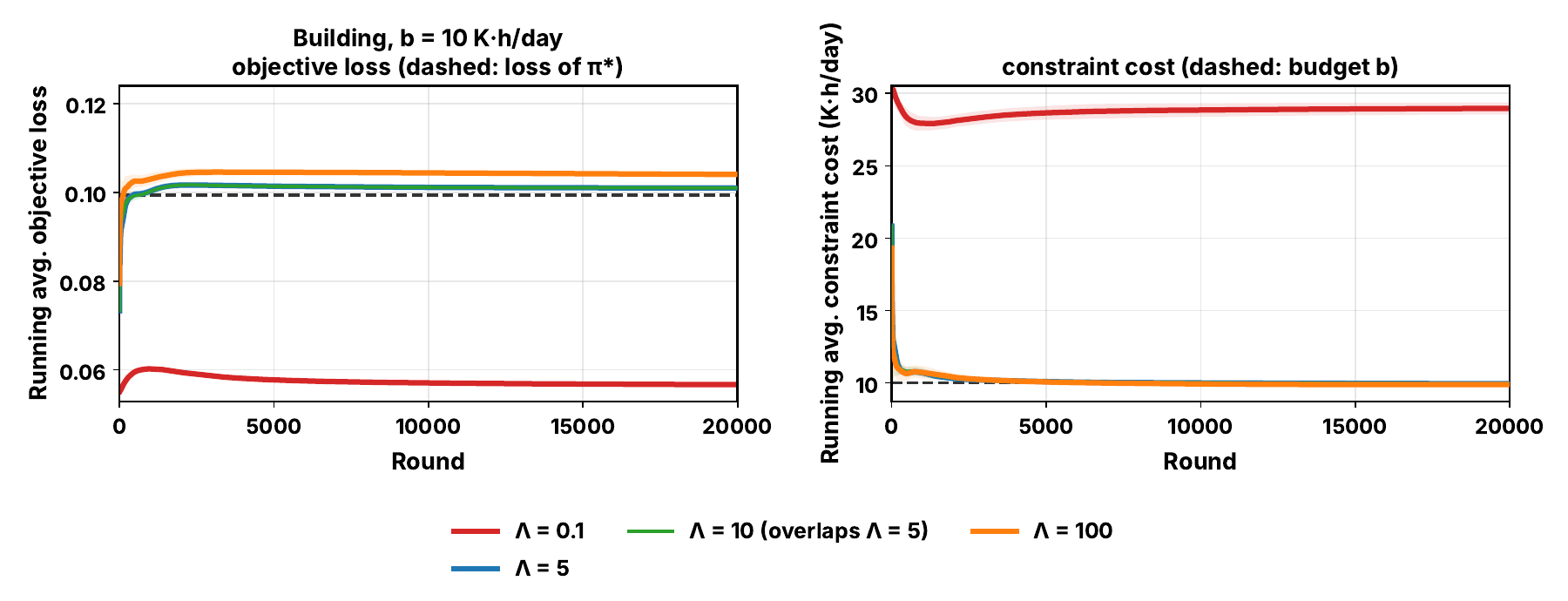}\\[2pt]
  \includegraphics[width=\linewidth]{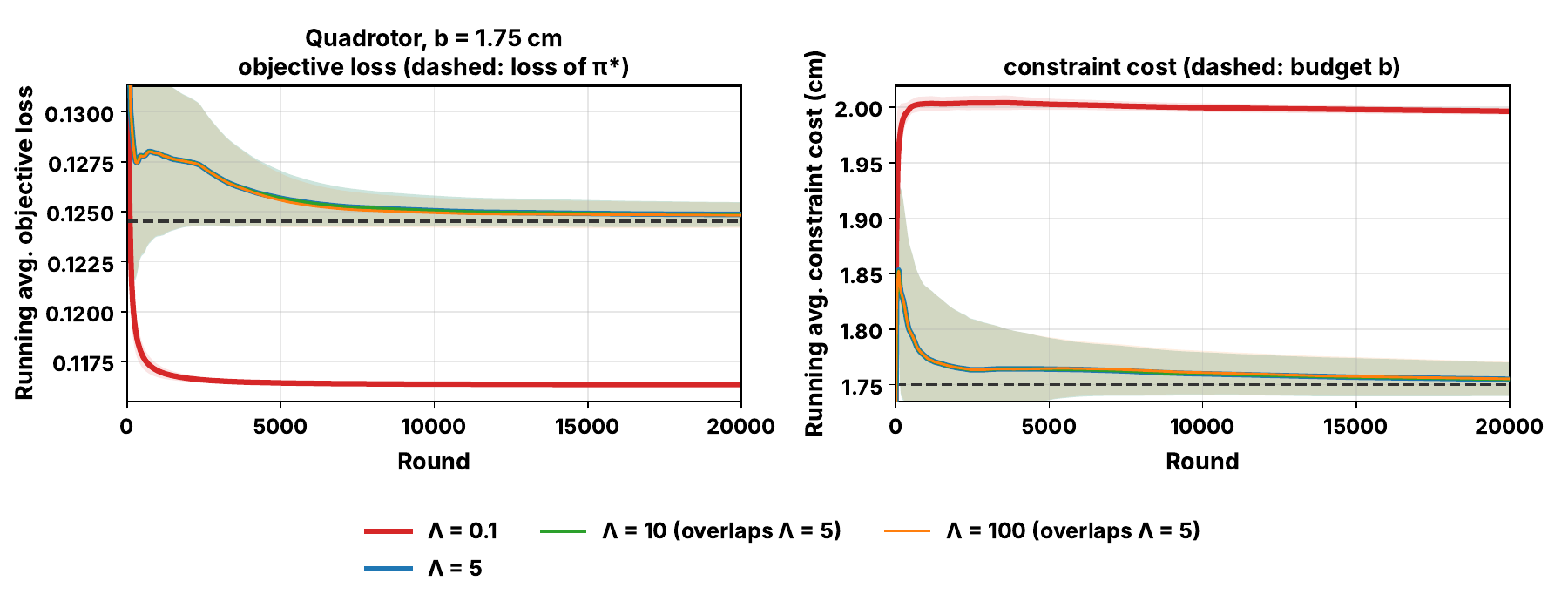}
  \caption{Running average of the expected loss (left) and cost (right) of our method for
  $\Lambda \in \{0.1, 5, 10, 100\}$, on the building with $b = 10$\,K$\cdot$h/day (top) and
  the quadrotor with $b = 1.75$\,cm (bottom). Dashed lines: loss of $\pi^*$ and budget $b$.
  Overlapping curves are drawn thinner on top. Mean $\pm 2$\,s.e.\ over 5 runs.}
  \label{fig:lambda}
\end{figure}

The clip level $\Lambda$ is the upper end of the interval $[0,\Lambda]$ in which the rectified weight $\mu_t$ is chosen (Definition~\ref{def:mu}). Our method does not tune it: it is fixed to $\Lambda = 2/\delta$, which is larger than $\lambda^\star$ because $\lambda^\star \le 1/\delta$ (Lemma~\ref{lem:lamstar}). The analysis uses $\Lambda$ in two places. The violation bound of Theorem~\ref{thm:viol} needs $\Lambda > \lambda^\star$: its proof divides by
$\Lambda - \lambda^\star \ge 1/\delta$. The regret bound of Theorem~\ref{thm:reward}, $R_T + 2\Lambda E_c$, grows with $\Lambda$. To see both effects, we repeat the main runs (same budgets, tuned $\gamma_0$, $T = 2\cdot10^4$, five runs)
and change only the clip level, $\Lambda \in \{0.1, 5, 10, 100\}$, with $\gamma$ scaled with $\Lambda$ as in Algorithm~1 so that $\Lambda = 2/\delta$ recovers the tuned run. Figure~\ref{fig:lambda} shows the running averages of the loss and of the cost.
\paragraph{Building ($b = 10$\,K$\cdot$h/day, $\lambda^\star = 0.31$).} With $\Lambda = 0.1 < \lambda^\star$ the weight stays at the clip and the discomfort stays near $29$\,K$\cdot$h/day,
almost three times the budget, for the whole run; the loss ($0.057$) is below that of $\pi^\star$ ($0.099$) only because the comfort constraint is ignored. With $\Lambda = 5$ and $10$ the two runs coincide: the discomfort settles very close to budget within about $2500$ rounds and the loss ($0.101$) is close to that of $\pi^\star$. With $\Lambda = 100$ the discomfort also stays close to budget, but the loss is slightly higher ($0.104$), because the larger $\Lambda$ gives a smaller $\gamma$ and hence more exploration. 
\paragraph{Quadrotor ($b = 1.75$\,cm, $\lambda^\star = 1.16$).} With $\Lambda = 0.1 < \lambda^\star$ the weight stays at the clip and the average overshoot stays near $2.0$\,cm,
above the budget; the loss ($0.116$) is again below that of $\pi^\star$ ($0.1245$) only because the constraint is ignored. The runs with $\Lambda = 5$, $10$ and $100$ coincide: the overshoot approaches the budget ($1.755$\,cm at the
end) and the loss approaches that of $\pi^\star$ ($0.125$).
\paragraph{Relation to the theory.}
The runs with $\Lambda = 0.1$ show what happens when the condition $\Lambda > \lambda^\star$ used in the proof of Theorem~\ref{thm:viol} fails: the cost stays above the budget on both benchmarks. Their low loss also illustrates the point of Section~\ref{sec:intro} that a small regret alone is not a meaningful guarantee, since it can be obtained by violating the constraint. For every $\Lambda \ge 5$, which exceeds $\lambda^\star$ on both benchmarks, the cost settles at the budget and the results hardly depend on $\Lambda$, so $\Lambda$ need not be tuned. A very large
$\Lambda$ remains safe for the constraint but may cost some loss, as on the building at $\Lambda = 100$, in line with Theorem~\ref{thm:reward}, whose regret bound grows with $\Lambda$. The default $\Lambda = 2/\delta$ ($18.0$ for the
building, $30.0$ for the quadrotor) lies in the region where the method works well, and by Lemma~\ref{lem:lamstar} choosing it requires no knowledge of $\lambda^\star$.

\section{Missing proofs for Lemma in Section \ref{sec:setup}}
\label{app:setup}
\subsection{Restatement and proof of Lemma \ref{lem:oracle}}
\begin{lemma}[(Restatement of Lemma \ref{lem:oracle})Consequences of the oracle bounds]
With probability at least $1-4\zeta$, for $g \in \{r, c\}$,
\begin{align*}
  \sum_{t=1}^T \mathbb{E}_{a \sim p_t(\cdot\mid x_t)}\big(\hat f^t_g - f^*_g\big)^2(x_t,a)
    &\le U_g := 2\,\mathrm{Reg}^g_{\mathsf{Sq}}(T) + 2\log(1/\zeta), \\
  \sum_{t=1}^T \big(V_c^{p_t} - \hat V_c^{p_t}\big)^2
    &\le U^{\mathcal{D}}_c := 2U_c + 2\log(1/\zeta).
\end{align*}
\end{lemma}

\begin{proof}
We use the following fact. 

Let $\mathcal{G}_0\subseteq\mathcal{G}_1\subseteq\cdots$ be a filtration
and $Y_t\in[0,1]$ be $\mathcal{G}_t$-measurable.

\begin{equation*}
\begin{aligned}
    &\text{Since }e^{-y}\le1-(1-e^{-1})y ~\text{on}~ [0,1],\\
    &\mathbb{E}[e^{-Y_t}\mid\mathcal{G}_{t-1}]\le\exp(-(1-e^{-1})\mathbb{E}[Y_t\mid\mathcal{G}_{t-1}]).
\end{aligned}
\end{equation*} 
so
$\exp\bigl(\sum_{s\le t}(1-e^{-1})\mathbb{E}[Y_s\mid\mathcal{G}_{s-1}]-Y_s\bigr)$ is a
supermartingale with mean at most $1$, and Markov's inequality gives, with probability at least
$1-\zeta$,
\begin{equation}\label{eq:freedman}
  \textstyle\sum_t\mathbb{E}[Y_t\mid\mathcal{G}_{t-1}]\le\frac{1}{1-e^{-1}}\bigl(\sum_tY_t+\log(1/\zeta)\bigr)
  \le2\sum_tY_t+2\log(1/\zeta).
\end{equation}
For the first bound, let $\mathcal{G}_{t-1}=\sigma(\mathcal{F}_{t-1},x_t)$ and
$Y_t=(\hat f^t_g-f^*_g)^2(x_t,a_t)$, which is $\mathcal{F}_t\subseteq\mathcal{G}_t$-measurable
with $\mathbb{E}[Y_t\mid\mathcal{G}_{t-1}]=\mathbb{E}_{a\sim p_t}(\hat f^t_g-f^*_g)^2(x_t,a)$;
combine \eqref{eq:freedman} with Assumption~\ref{ass:oracle}. For the second bound, Jensen's
inequality gives
$(V_c^{p_t} - \hat V_c^{p_t})^2 \le \mathbb{E}_{x\sim\mathcal{D}}\mathbb{E}_{a\sim p_t}(\hat f^t_c - f^*_c)^2(x,a)$.
This is $\mathbb{E}[Y'_t\mid\mathcal{F}_{t-1}]$ for
$Y'_t=\mathbb{E}_{a\sim p_t}(\hat f^t_c - f^*_c)^2(x_t,a)$, because $x_t \sim \mathcal{D}$ is
independent of $\mathcal{F}_{t-1}$ and $p_t$, $\hat f^t_c$ are $\mathcal{F}_{t-1}$-measurable;
apply \eqref{eq:freedman} with $\mathcal{G}_t=\mathcal{F}_t$ and then the first bound. The four
events (oracle, and \eqref{eq:freedman} three times) fail with total probability at most $4\zeta$.
\end{proof}

\section{Proofs for Lemmas in Section \ref{sec:algo}}
\label{app:algo}
\subsection{Restatement and Proof for Lemma \ref{lem:shape}}

\begin{lemma}[(Restatement of Lemma \ref{lem:shape}) Shape of $g_t$ ]
$g_t$ is nonincreasing and right-continuous on $[0,\infty)$, and all its discontinuities are
downward jumps.
\end{lemma}

\begin{proof}
Fix a context $x$ and write $r_a=\hat f_r^t(x,a)$, $c_a=\hat f_c^t(x,a)$, $z_a(\mu)=r_a+\mu c_a$
and $h(\mu)=\sum_a p^{(\mu)}_{t,a}c_a$ for the predicted cost at $x$. 

Each score $z_a$ is a line in $\mu$ with slope $c_a$, and the leader is the lowest line. Since there are finitely many lines, $[0,\infty)$ splits into finitely many intervals on which the leader $\ell$ is fixed, separated by points where two or more lines cross at the bottom.

\emph{\textbf{Step 1: $h$ is non-increasing while the leader is fixed.}} The leader's probability is
$1-\sum_{a\ne\ell}p^{(\mu)}_{t,a}$, so
$h(\mu)=c_\ell+\sum_{a\ne\ell}p^{(\mu)}_{t,a}(c_a-c_\ell)$. The gap
$z_a-z_\ell$ has derivative $c_a-c_\ell$, so
$p^{(\mu)}_{t,a}=1/(K+\gamma(z_a-z_\ell))$ has derivative
$-\gamma(p^{(\mu)}_{t,a})^2(c_a-c_\ell)$, and
\[
  h'(\mu)=-\gamma\sum_{a\ne\ell}(p^{(\mu)}_{t,a})^2(c_a-c_\ell)^2\le0 .
\]
In words, increasing $\mu$ moves probability from actions that cost more than the leader to
the leader, and from the leader to actions that cost less; both lower the cost.

\emph{\textbf{Step 2: what happens where the leader changes.}} Let $\mu^\circ$ be a point where the
leader changes, from $\ell$ just below $\mu^\circ$ to $\ell'$ just above. Both lines are lowest
at $\mu^\circ$, and $\ell$ was below $\ell'$ before, so $\ell$'s line is steeper:
$c_\ell>c_{\ell'}$. At $\mu^\circ$ itself the two scores are equal, so either could be called
the leader; the tie-break picks the one with the smaller cost, $\ell'$, which is also the
leader just above $\mu^\circ$. Hence $h(\mu^\circ)=h(\mu^\circ{+})$: $h$ is right-continuous.

\emph{\textbf{Step 3: the jump is downward.}} Compare the distributions just below and at $\mu^\circ$.
They are inverse-gap weighting for the same scores $z(\mu^\circ)$ (in the limit), differing only in
which of the tied actions is called the leader. Every action other than $\ell,\ell'$ has the
same gap, hence the same probability, in both. The tied action that is not the leader has gap
$0$, hence probability $1/K$. The leader gets the remaining mass $P$, which is the same in both.
So just below $\mu^\circ$, $\ell$ has probability $P$ and $\ell'$ has $1/K$; at $\mu^\circ$ these
are swapped, and
\[
  h(\mu^\circ)-h(\mu^\circ{-})=\bigl(Pc_{\ell'}+\tfrac1Kc_\ell\bigr)-\bigl(Pc_\ell+\tfrac1Kc_{\ell'}\bigr)
  =(c_{\ell'}-c_\ell)\bigl(P-\tfrac1K\bigr)\le0,
\]
because $c_{\ell'}<c_\ell$, and $P\ge1/K$ since each of the $K-1$ non-leaders has probability
at most $1/K$. If more than two actions tie, the same computation applies with the extra tied
actions at probability $1/K$ on both sides.

\emph{\textbf{Averaging over contexts.}} Steps 1--3 show that $h$ is nonincreasing and
right-continuous for each $x$. Both properties are preserved by averaging over
$x\sim\mathcal{D}$ (for right-continuity, by bounded convergence), and a nonincreasing function
can only jump downward.
\end{proof}

\subsection{Proof for Proposition \ref{prop:fp}}
\label{app:prop_fp}

\begin{proposition}[(Restatement of Proposition \ref{prop:fp}) Existence, computability]
The policy $p_t$ of Definition~\ref{def:mu} is well defined, and in the third case
$\hat V_c^{p_t} = b$. It can be computed by bisection on $[0,\Lambda]$. In every case $p_t$ is,
for each $x$, a convex combination of inverse-gap-weighted distributions for the score
vector $\hat y_t(\mu_t)$, each with a minimizer of $\hat y_t(\mu_t)$ as the leader. Finally,
$\mu_t$ and $p_t$ depend only on $\hat f_r^t, \hat f_c^t$, hence are
$\mathcal F_{t-1}$-measurable; in particular they do not depend on the realized action $a_t$.
\end{proposition}

\begin{proof}
In the third case $g_t(0) > b \ge g_t(\Lambda)$, so $\mu_t\in(0,\Lambda]$ exists. By
Lemma~\ref{lem:shape}, $g_t(\mu_t) \le b$ (right-continuity) and $\hat V_c^{\bar p^{(\mu_t)}_t} = g_t(\mu_t{-})
\ge b$ (Step~3 of Lemma~\ref{lem:shape} and the definition of the infimum), so a $w_t$ with $\hat V_c^{p_t} = b$ exists. Bisection keeps
a point $\mu^{\mathrm{lo}}$ with $g_t > b$ and a point $\mu^{\mathrm{hi}}$ with $g_t \le b$ and
mixes the two policies; after $n$ steps $\mu^{\mathrm{hi}} - \mu^{\mathrm{lo}} \le \Lambda 2^{-n}$,
and every bound below holds up to an additive $T\Lambda 2^{-n}$. Both $\bar p^{(\mu_t)}_t$ and
$p^{(\mu_t)}_t$ are inverse-gap weighting for $\hat y_t(\mu_t)$ with a minimizer as the leader.
Everything in $g_t$ is a function of $\hat f_r^t,\hat f_c^t$, which are determined by data from
rounds $<t$.
\end{proof}

\section{Missing proofs for Lemmas in Section \ref{sec:scb}}
\label{app:scb}
\textbf{Proof of Lemma~\ref{lem:igw}}: 
\begin{proof}
Let $\ell$ be the leader, $a^\star\in\arg\min_af_a$ and $\Delta_a=\hat y_a-\hat y_\ell\ge0$. Then
\[
  \textstyle\sum_ap_af_a-f_{a^\star}
  =\sum_{a\ne a^\star}p_a(f_a-\hat y_a)+(1-p_{a^\star})(\hat y_{a^\star}-f_{a^\star})
   +\sum_ap_a\Delta_a-\Delta_{a^\star}.
\]
Maximizing $p_au-\tfrac\gamma4p_au^2$ and $(1-p_{a^\star})u-\tfrac\gamma4p_{a^\star}u^2$ over
$u\in\mathbb{R}$ bounds the first two terms, minus $\tfrac\gamma4\sum_ap_a(\hat y_a-f_a)^2$, by
$\sum_{a\ne a^\star}\tfrac{p_a}{\gamma}+\tfrac{(1-p_{a^\star})^2}{\gamma p_{a^\star}}
=\tfrac{1-p_{a^\star}}{\gamma p_{a^\star}}\le\tfrac{1}{\gamma p_{a^\star}}$. Next,
$\sum_ap_a\Delta_a=\sum_{a\ne\ell}\tfrac{\Delta_a}{K+\gamma\Delta_a}\le\tfrac{K-1}{\gamma}$. If
$a^\star\ne\ell$, then $\tfrac1{\gamma p_{a^\star}}=\tfrac K\gamma+\Delta_{a^\star}$; if
$a^\star=\ell$, then $p_{a^\star}\ge1/K$ and $\Delta_{a^\star}=0$. In both cases the total is at
most $2K/\gamma$.
\end{proof}
\begin{lemma}[(Restatement of Lemma \ref{lem:sqcb}) Composite regret]
For any fixed comparator policy $\pi$, with probability at least $1-5\zeta$,
\begin{equation}\label{eq:sqcb}
  \sum_{t=1}^T\Bigl[(V_r^{p_t}+\mu_t V_c^{p_t})-(V_r^{\pi}+\mu_t V_c^{\pi})\Bigr]
  \le R_T:=2\sqrt{KT\,(U_r+\Lambda^2U_c)} + (1+\Lambda)\sqrt{2T\log(1/\zeta)} .
\end{equation}
With $\Lambda=2/\delta$, $R_T=O\bigl(\sqrt{KT\,\mathrm{Reg}_{\mathsf{Sq}}(T)}/\delta\bigr)$.
\end{lemma}

\begin{proof}
At round $t$ let $f_t = f_r^*(x_t,\cdot)+\mu_t f_c^*(x_t,\cdot)$, the conditional mean of the
composite loss $\ell_t + \mu_t c_t$; it is linear in the observable means, so there is no
Jensen gap, and $\hat y_t(\mu_t)$ is its estimate. Apply \eqref{eq:lem3} at the realized
context $x_t$, with $\min_a f_{t,a} \le \mathbb{E}_{a\sim\pi(\cdot\mid x_t)} f_{t,a}$; this is
valid even though $f_t$ changes with $t$. Since
$(\hat y_{t,a} - f_{t,a})^2 \le 2(\hat f^t_r - f^*_r)^2 + 2\Lambda^2(\hat f^t_c - f^*_c)^2$, summing
and using Lemma~\ref{lem:oracle} gives
\[
  \sum_t \Bigl[\mathbb{E}_{a\sim p_t(\cdot\mid x_t)} f_{t,a} - \mathbb{E}_{a\sim\pi(\cdot\mid x_t)} f_{t,a}\Bigr]
  \le \frac{2KT}{\gamma} + \frac{\gamma}{2}(U_r + \Lambda^2U_c) = 2\sqrt{KT(U_r+\Lambda^2U_c)}
\]
for our $\gamma$. Since $p_t$ and $\mu_t$ are $\mathcal{F}_{t-1}$-measurable and
$x_t\sim\mathcal{D}$ is independent of $\mathcal{F}_{t-1}$, the conditional mean of the $t$-th
summand is $(V_r^{p_t}+\mu_t V_c^{p_t})-(V_r^{\pi}+\mu_t V_c^{\pi})$. The differences lie in
$[-(1+\Lambda), 1+\Lambda]$, and the Azuma--Hoeffding inequality gives the last term.
\end{proof}

\section{Missing proofs in Section \ref{sec:mi2b}}
\label{app:mi2b}
\begin{lemma}[(Restatement of Lemma \ref{lem:extract}) Rectified extraction]
With $\mu_t$ and $p_t$ from Definition~\ref{def:mu},
\begin{equation}\label{eq:extract}
  \sum_{t=1}^T\mu_t\bigl(V_c^{p_t}-b\bigr)\;\ge\;\Lambda\,\mathrm{Viol}(T)-2\Lambda E_c .
\end{equation}
\end{lemma}

\begin{proof}
Fix $t$ and let $\epsilon_t=|V_c^{p_t}-\hat V_c^{p_t}|$. We show
$\mu_t(V_c^{p_t}-b)\ge\Lambda(V_c^{p_t}-b)^+-2\Lambda\epsilon_t$ in each
case of Definition~\ref{def:mu}.

\emph{Case $\mu_t=0$ ($\hat V_c^{p_t}=g_t(0)\le b$).} The left side is $0$.
Since $\hat V_c^{p_t}\le b$, $(V_c^{p_t}-b)^+\le\epsilon_t$, so
$\Lambda(V_c^{p_t}-b)^+-2\Lambda\epsilon_t\le0$.

\emph{Case $\mu_t=\Lambda$ with $\hat V_c^{p_t}=g_t(\Lambda)>b$.}
The left side is $\Lambda(V_c^{p_t}-b)$, which differs from $\Lambda(V_c^{p_t}-b)^+$ by
$\Lambda\min(V_c^{p_t}-b,0)\ge-\Lambda\epsilon_t$, since $V_c^{p_t}>b-\epsilon_t$.

\emph{Third case ($\hat V_c^{p_t}=b$, $\mu_t\in(0,\Lambda]$).}
Then $|V_c^{p_t}-b|=\epsilon_t$, so the left side is at least $-\mu_t\epsilon_t\ge-\Lambda\epsilon_t$,
while $\Lambda(V_c^{p_t}-b)^+-2\Lambda\epsilon_t\le-\Lambda\epsilon_t$.

Summing over $t$ and using $\sum_t\epsilon_t\le E_c$ gives
\eqref{eq:extract}.
\end{proof}

\section{The Exact-Penalty Saddle Inequality}

\begin{lemma}[Bounded optimal multiplier]\label{lem:lamstar}
Under Assumption~\ref{ass:slater} there is an optimal Lagrange
multiplier $\lambda^*\ge0$ for
$\min\{V_r^\pi:V_c^\pi\le b\}$ with $\lambda^*\le1/\delta$, satisfying
$V_r^\pi+\lambda^*(V_c^\pi-b)\ge V_r^{\pi^*}$ for all randomized $\pi$
and $\lambda^*(V_c^{\pi^*}-b)=0$.
\end{lemma}

\begin{proof}
$V_r,V_c$ are linear over the convex set of randomized policies and
$\pi^s$ is strictly feasible, so strong duality holds and an optimal
$\lambda^*$ exists with the stated optimality and complementary
slackness. Evaluating Lagrangian optimality at $\pi^s$:
$V_r^{\pi^*}\le V_r^{\pi^s}+\lambda^*(V_c^{\pi^s}-b)\le V_r^{\pi^s}-\lambda^*\delta$,
so $\lambda^*\le(V_r^{\pi^s}-V_r^{\pi^*})/\delta\le1/\delta$
(values in $[0,1]$).
\end{proof}

\begin{lemma}[Quantitative saddle]\label{lem:qsaddle}
For every randomized $\pi$,
\(
  V_r^\pi-V_r^{\pi^*}\ge-\lambda^*(V_c^\pi-b)^+ .
\)
\end{lemma}

\begin{proof}
By Lemma~\ref{lem:lamstar},
$V_r^\pi-V_r^{\pi^*}\ge-\lambda^*(V_c^\pi-b)$. Since $\lambda^*\ge0$
and $x\le(x)^+$, $-\lambda^*(V_c^\pi-b)\ge-\lambda^*(V_c^\pi-b)^+$.
\end{proof}

\section{Hyperparameters and Primal-Dual Square CB}
\begin{table}[h]
\floatconts
  {tab:hyper}
  {\caption{Parameters of all methods in the experiments. The algorithm sets
   $\gamma=2\sqrt{KT/(U_r+\Lambda^2U_c)}$, but the oracle bounds $U_r,U_c$ are unknown in practice. Primal-dual uses the dual step $\eta=\eta_0\Lambda/\sqrt{T}$. Tuned on separate seeds by the smallest mean of regret plus violation, over logarithmic
   grids.}}
  {\begin{tabular}{llccc}
    \toprule
    Parameter & Description & Tuning grid & Quadrotor & Building \\
    \midrule
    $\Lambda$ & clip level, $\Lambda = 2/\delta$              & --- (not tuned)    & 30 & 18 \\
    $T$       & horizon (number of episodes)                  & ---                & $200{,}000$ & $50{,}000$ \\
    $K$       & number of actions                             & ---                & 6 & 8 \\
    $b$       & budget                                        & ---                & 1.75\,cm & 10\,K$\cdot$h/day \\
    $\delta$  & Slater margin, $\delta = b - V_c^{\min}$      & ---                & 0.0667 & 0.1111 \\
    \midrule
    $\gamma_0$ & ours: exploration                            & $[3,\,10^5]$       & $10^{2}$ & $10^{1}$ \\
    $\gamma_0$ & primal-dual: exploration                     & $[3,\,10^5]$       & $10^{5}$ & $10^{5}$ \\
    $\eta_0$   & primal-dual: dual step                       & $[0.01,\,10]$      & $0.3$ & $0.03$ \\
    $U_V$      & CCB method: queue scale                      & $[10^{-4},\,10^{3}]$ & $0.01$ & $1$ \\
    $U_\gamma$ & CCB method: exploration scale                & $[10^{-8},\,10]$   & $10^{-6}$ & $10^{-6}$ \\
    \bottomrule
  \label{tab:hyp}
  \end{tabular}}
\end{table}

\begin{center}
\fbox{\parbox{0.93\textwidth}{
\textbf{Algorithm: Primal-dual SquareCB (baseline).}\\
Parameters $\gamma$, step size $\eta$, budget $b$; oracles $\mathsf{SqAlg}_r,\mathsf{SqAlg}_c$;
$\mu_{1}=0$.\\[3pt]
For $t=1,\dots,T$:
\begin{enumerate}\itemsep1pt
  \item Receive $x_t$; obtain $\hat f_r^t(x_t,\cdot),\hat f_c^t(x_t,\cdot)$.
  \item Form scores $\hat y_{t,a}=\hat f_r^t(x_t,a)+\mu_t\hat f_c^t(x_t,a)$
        and the SquareCB distribution $p_t=p^{(\mu_t)}_t$ via \eqref{eq:p-of-mu}.
  \item Compute the estimated cost $\hat c_t = \sum_a p_t(a\mid x_t)\,\hat f_c^t(x_t,a)$.
  \item Update $\mu_{t+1} = \max\{\mu_{t} + \eta (\hat c_t-b),0\}$.
  \item Sample $a_t\sim p_t$; observe $\ell_t(a_t),c_t(a_t)$; update both oracles.
\end{enumerate}
}}
\end{center}

\end{document}